\documentclass[10pt,a4paper,twocolumn]{article}

\usepackage[utf8]{inputenc}
\usepackage{authblk}
\usepackage{graphicx}
\usepackage[explicit]{titlesec}
\usepackage{color, colortbl}
\usepackage{url}
\usepackage[hang]{footmisc}
\usepackage[normalem]{ulem}
\usepackage[top=2.5cm, bottom=2.8cm, left=1.5cm, right=1.5cm]{geometry}
\usepackage{abstract}
\usepackage{balance}
\usepackage{makecell}

\providecommand{\keywords}[1]{\textbf{Keywords}\ \ \textendash\ \   #1}

\titleformat{\section}{\large\bfseries}{\thesection.}{1em}{\MakeUppercase{#1}}
\titlespacing*{\section}{0pt}{12pt}{6pt}
\titleformat{\subsection}{\large}{\thesubsection}{1em}{#1}
\titlespacing*{\subsection}{0pt}{12pt}{6pt}
\titleformat{\subsubsection}{\large\itshape}{\thesubsubsection}{1em}{#1}
\titlespacing*{\subsubsection}{0pt}{12pt}{6pt}
\newcommand{\ITUpar}{\vspace{8pt}\par}
\renewenvironment{abstract}
               {\list{}{
               \setlength{\rightmargin}{0mm}
               \setlength{\leftmargin}{0mm}
               \vspace{-0.25in}
                \item[\textit{\textbf{\hspace{22pt}Abstract  }}  \textendash]\relax}}
               {\endlist}

\definecolor{Gray}{gray}{0.97}
\newcommand\blfootnote[1]{%
  \begingroup
  \renewcommand\thefootnote{}\footnote{#1}%
  \addtocounter{footnote}{-1}%
  \endgroup
}

\title{\large{\textbf{\uppercase{Trends and Advancements in\\ Deep Neural Network Communication}}}}

\author[1]{\normalsize{Felix Sattler}}
\author[1,2]{\normalsize{Thomas Wiegand}}
\author[1]{\normalsize{Wojciech Samek}}

\affil[1]{\normalsize{Department of Video Coding \& Analytics, Fraunhofer Heinrich Hertz Institute, 10587 Berlin, Germany}}
\affil[2]{\normalsize{Department of Electrical Engineering \& Computer Science, Technische Universit\"at Berlin, 10587 Berlin, Germany}}

\date{\vspace{-12pt}\endgraf\rule{\textwidth}{1pt}}

\begin{document}
\twocolumn[

\begin{@twocolumnfalse}
\maketitle

\begin{abstract}
\textit{
Due to their great performance and scalability properties neural networks have become ubiquitous building blocks of many applications. With the rise of mobile and IoT, these models now are also being increasingly applied in distributed settings, where the owners of the data are separated by limited communication channels and privacy constraints. To address the challenges of these distributed environments, a wide range of training and evaluation schemes have been developed, which require the communication of neural network parametrizations. These novel approaches, which bring the "intelligence to the data" have many advantages over traditional cloud solutions such as privacy-preservation, increased security and device autonomy, communication efficiency and high training speed. This paper gives an overview over the recent advancements and challenges in this new field of research at the intersection of machine learning and communications.
}
\end{abstract}

\ITUpar
\keywords{Neural networks, federated learning, model compression, distributed training, on-device inference.}

\ITUpar
\ITUpar

\end{@twocolumnfalse}
]

\section{Introduction}
\blfootnote{This work was partly supported by the German Ministry for Education and Research as BIFOLD - Berlin Institute for the Foundations of Learning and Data (ref. 01IS18025A and ref 01IS18037A).}
Neural networks have achieved impressive successes in a wide variety of areas of computational intelligence such as computer vision \cite{hinton2012deep}\cite{xu2015show}\cite{karpathy2015deep}, natural language processing \cite{bahdanau2014neural}\cite{kim2016character}\cite{sutskever2014sequence} and speech recognition \cite{graves2005framewise} among many others and, as a result, have become a core building block of many applications. 

As mobile and internet of things (IoT) devices become ubiquitous parts of our daily lives, neural networks are also being applied in more and more distributed settings. These distributed devices are getting equipped with ever more potent sensors and storage capacities and collect vast amounts of personalized data, which is highly valuable for processing in machine learning pipelines.

When it comes to processing of data from distributed sources, the \emph{"Cloud ML"} paradigm \cite{hwang2017cloud} has reigned supreme in the previous decade. 
In Cloud ML, local user data is communicated from the often hardware constrained mobile or IoT devices to a computationally potent centralized server where it is then processed in a machine learning pipeline (e.g. a prediction is made using an existing model or the data is used to train a new model). The result of the processing operation may then be sent back to the local device. 
From a communication perspective, methods which follow the Cloud ML paradigm make use of centralized intelligence and 
\begin{center}
\textit{
"Bring the data to the model."}
\end{center} 

\begin{figure*}
\centering
\includegraphics[width=0.95\textwidth]{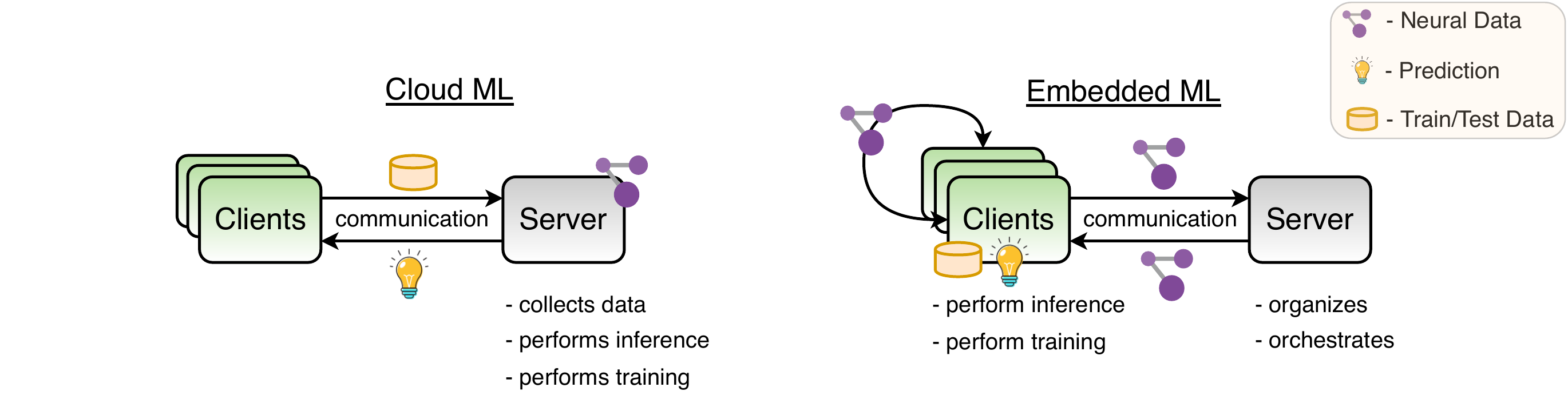}
\caption{Comparison between the two paradigms for machine learning from distributed data. In cloud ml, data from users is collected and processed by a centralized service provider. In Embedded ML, data never leaves the user device. To perform inference and collaborative training,  neural network parametrizations are communicated and data is processed locally.  }
\end{figure*}
While the Cloud ML paradigm is convenient for the clients from a computational perspective, as it moves all the workload for processing the data to the computationally potent server, it also has multiple severe drawbacks and limitations, which all arise from the fact that user data is processed at a centralized location:\\

{\bf Privacy}: Data collected by mobile or IoT devices is often of private nature and thus bound to the local device. Medical data, text messages, private pictures or footage from surveillance cameras are examples for data which can not be processed in the cloud. New data protection legislations like the European GDPR \cite{voigt2017eu} or the Cyber Security Law of the People's Republic of China enforce strong regulations on data privacy. 

{\bf Ownership}: Attributing and claiming ownership is a difficult task if personal data is transfered to a central location. Cloud ML leaves users in the dark about what happens with their data or requires cumbersome rights management from the cloud service provider.

{\bf Security}: With all data being stored at one central location, Cloud ML exposes a single point of failure. Multiple cases of data leakage in recent times\footnote{A comprehensive list of documented breaches can be found at \url{https://en.wikipedia.org/wiki/List_of_data_breaches} .} have demonstrated that the centralized processing of data comes with an unpredictable security risk for the users.

{\bf Efficiency}: Transferring large records of data to a central compute node often is more expensive in terms of time and energy than the actual processing of the data. For instance, single records of medical image data can already be hundreds of Megabytes in size \cite{varma2012managing}. If the local data is large and/or the communication channels are limited, moving data to the cloud might thus become inefficient or unfeasible. 

{\bf Autonomy}: Many distributed devices need to act fully autonomously and are not allowed to depend on slow and unreliable connections to a cloud server. For instance, in a self-driving car, intelligence responsible for making driving decisions needs to be available at all times and thus has to be present on-device. 

\ \\
As awareness for these issues increases and mobile and IoT devices are getting equipped with ever more potent hardware, a new paradigm, which we term \emph{"Embedded ML"}, arises with the goal to keep data on device and
\begin{center}
\textit{
"Bring the model to the data."}
\end{center}

Multi-party machine learning workflows that follow this paradigm all have one principle in common: In order to avoid the shortcomings of Cloud ML and achieve data locality they communicate neural network parametrizations ("neural data") instead of raw data. This may include not only trained neural network models, but also model updates and model gradients. 

Since neural networks are typically very large, containing millions to billions of parameters \cite{shoeybi2019megatron}, and mobile connections are slow, unreliable and costly the communication of neural data is typically one of the main bottlenecks in applications of Embedded ML. As a result, recently a vast amount of research has been conducted, which aims to reduce the size of neural network representations and a wide range of domain specific compression methods have been proposed. 

In this work, we provide an overview on machine learning workflows which follow the Embedded ML paradigm through the unified lens of communication efficiency. We describe properties of the "neural data" communicated in Embedded ML and systematically review the current state of research in neural data compression. Finally, we also enumerate important related challenges, which need to be considered when designing efficient communication schemes for Embedded ML applications.

\section{Survey on Neural Network Communication}
\begin{figure*}[t]
\centering
\includegraphics[width=\textwidth]{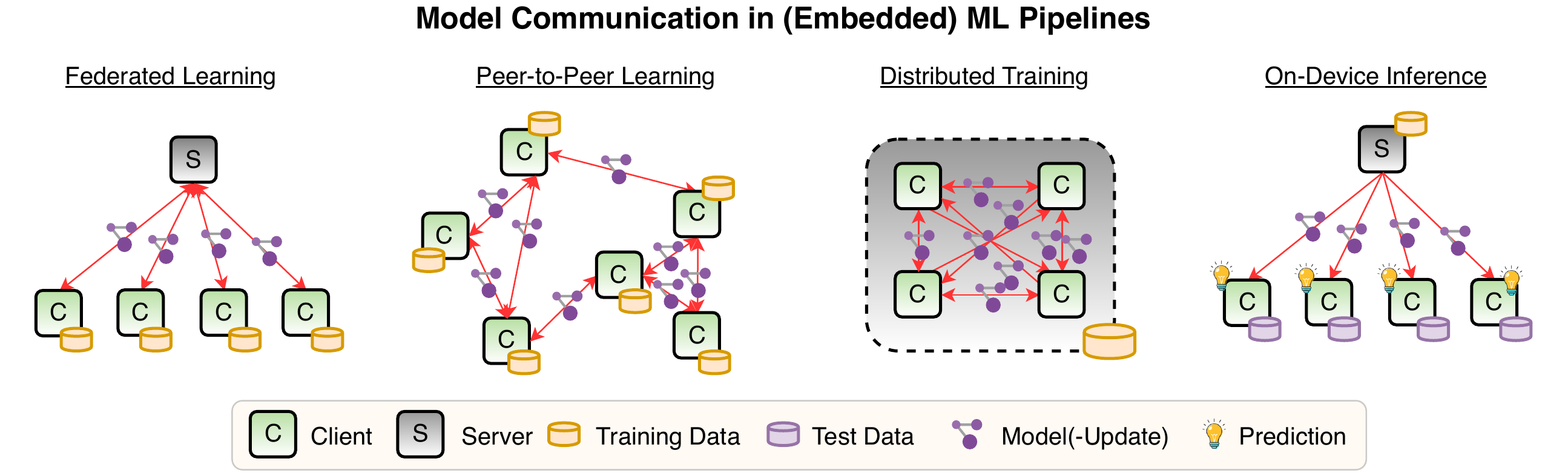}
\caption{Model communication at the training and inference stages of different Embedded ML pipelines. 
From left to right: (1) Federated learning allows multiple clients to jointly train a neural network on their combined data, without any of the local clients having to compromise the privacy of their data. This is achieved by iteratively exchanging model updates with a centralized server. (2) In scenarios where it is undesirable to have a centralized entity coordinating the collaborative training process, \emph{peer-to-peer learning} offers a potential solution. In peer-to-peer learning the clients directly exchange parameter updates with their neighbors according to some graph predefined topology. (3) In the data center setting, training speed can be drastically increased by splitting the workload among multiple training devices via distributed training. This however requires frequent communication of model gradients between the learner devices. (4) On-device inference protects user privacy and allows fast and autonomous predictions, but comes at the cost of communicating trained models from the server to the individual users.}
\label{fig:2}
\end{figure*}

We currently witness the emergence of a variety of applications of Embedded ML, where neural networks are being communicated.
In this section we will review the three most important settings, namely on-device inference, federated learning and peer-to-peer learning. These settings differ with respect to their communication topology, frequency of communication and network constraints. We will also review distributed training in the data center, as many methods for neural data compression have been proposed in this domain. Figure \ref{fig:2} illustrates the flow of (neural) data in these different settings. Table \ref{tab:1} summarizes the communication characteristics of the different distributed ML pipelines in further detail and gives an overview on popular compression techniques in the respective applications.

\subsection{On-device Inference}
Inference is the act of using a statistical model (e.g. a trained neural network) to make predictions on new data. While cloud-based inference solutions can certainly offer a variety of benefits, there still exists a wide range of applications that require quick, autonomous and failure-proof decision making, which can only be offered by on-device intelligence solutions.

For instance, in a self-driving car, intelligence responsible for making driving decisions needs to be available at all times and thus has to be present on-device. At the same time, the models used for inference might be continuously improving as new training data becomes available and thus need to be frequently communicated from the compute node to a potentially very large number of user devices. Since typical modern DNNs consists of exorbitant numbers of parameters this constant streaming of models can impose a high burden on the communication channel, potentially resulting in prohibitive delays and energy spendings. 

\textbf{Compression for On-Device Inference:} The field of neural network compression has set out to mitigate this problem by reducing the size of trained neural network representations. The goal in this setting is typically to find a compressed neural network representation with minimal bit-size, which achieves the same or comparable performance as the uncompressed representation. To achieve this end, a large variety of methods have been proposed which vary w.r.t. the computational effort of encoding and compression results. We want to stress that neural network compression is a very active field of research and considers issues of communication efficiency, alongside other factors such as memory- and computation complexity, energy efficiency and specialized hardware. While we only focus on the communication aspect of neural network compression, a more comprehensive survey can be found e.g. in \cite{choudharycomprehensive}.

In neural network compression it is usually assumed that the sender of the neural network has access to the entire training data and sufficient computational resources to retrain the model. By using training data during the compression process the harmful effects of compression can be alleviated. The three most popular methods for trained compression are Pruning, Distillation and trained quantization.

Pruning techniques \cite{karnin1990simple}\cite{castellano1997iterative}\cite{han2015learning}\cite{yang2017designing} aim to reduce the entropy of the neural network representation by forcing a large number of elements to zero. This is achieved by modifying the training objective in order to promote sparsity. This is typically done by adding a $\ell_1$ or $\ell_2$ regularization penalty to the weights, but also Bayesian approaches \cite{molchanov2017variational} have been proposed. Pruning techniques have been shown to be able to achieve compression rates of ore than one order of magnitude, depending on the degree of overparameterization in the network \cite{han2015learning}. 

Distillation methods \cite{hinton2015distilling} can be used to transfer the knowledge of a larger model into a considerably smaller architecture. This is achieved by using the predictions of the larger network as soft-labels for the smaller network. 

Trained quantization methods restrict the bitwidth of the neural network during training, e.g., reducing the precision from 32 bit to 8 bit \cite{wang2018training}. Other methods generalize this idea and aim to directly minimize the entropy of the neural network representation during training \cite{wiedemann2019entropy}. It is important to note however, that all of these methods require re-training of the network and are thus computationally expensive and can only be applied if the full training data is available.

In situations where compression needs to be fast and/or no training data is available at the sending node, trained compression techniques can not be applied and one has to resort to ordinary lossy compression methods. Among these, (vector) quantization methods \cite{choi2018universal}\cite{choi2016towards} and efficient matrix decompositions \cite{tjandra2018tensor}\cite{yu2017compressing} are popular. 

A middle-ground between trained and ordinary lossy compression methods are methods which only require some data to guide the compression process. These approaches use different relevance measures based e.g. the diagonal of the Hessian \cite{hassibi1993second}, Fisher information \cite{tu2016reducing} or layer-wise relevance \cite{yeom2019pruning}\cite{bach2015pixel} to determine which parameters of the network are important.

Many of the above described techniques are somewhat orthogonal and can be combined. For instance the seminal "Deep Compression" method \cite{han2015deep} combines pruning with learned quantization and Huffman coding to achieve compression rates of up to x49 on a popular VGG model, without any loss in accuracy. More recently the DeepCABAC \cite{wiedemann2019deepcabac} algorithm, developed within the MPEG standardization initiative on neural network compression\footnote{https://mpeg.chiariglione.org/standards/mpeg-7/compression-neural-networks-multimedia-content-description-and-analysis}, makes use of learned quantization and the very efficient CABAC encoder \cite{marpe2003highly} to further increase the compression rate to x63.6 on the same architecture.

\begin{table*}[!ht]
\centering
\caption{Communication characteristics of different Embedded ML pipelines and popular respective compression techniques used in the literature to reduce communication.}
\label{tab:1}
\begin{tabular}{lcccc}
 & \makecell{\underline{On-Device}\\\underline{Inference}} & \makecell{\underline{Distributed}\\\underline{Training}} & \makecell{\underline{Federated}\\\underline{Learning}} & \makecell{\underline{Peer-to-Peer}\\\makecell{\underline{Learning}}} \\
\underline{Communication:}\\
\rowcolor{Gray}
$\bullet$ Objects & \makecell{trained models/\\model updates} & \makecell{model gradients} & \makecell{models/\\model updates} & \makecell{models/\\model updates}\\
$\bullet$ Flow & \makecell{server\\$\rightarrow$ clients} & \makecell{all clients\\$\rightarrow$ all clients} & \makecell{some clients\\$\leftrightarrow$ server} &\makecell{all clients\\$\rightarrow$ some clients} \\
\rowcolor{Gray}
$\bullet$ Frequency & low & high & medium & high \\
$\bullet$ Redundancy & low & high & medium & low\\
\\
\underline{Compression Techniques:}\\[+3pt]
\multicolumn{4}{l}{$\bullet$ Trained Compression:}\\
\rowcolor{Gray}
\hspace{0.5cm}$\rightarrow$ Pruning & \cite{han2015learning}\cite{wang2016deep}\cite{yang2017designing} & - & \cite{konevcny2016federated} & - \\
\hspace{0.5cm}$\rightarrow$ Trained Quantization & \cite{wang2018training}\cite{han2015learning}\cite{wiedemann2019entropy} & - & \cite{konevcny2016federated} & -\\
\rowcolor{Gray}
\hspace{0.5cm}$\rightarrow$ Distillation & \cite{hinton2015distilling} & - & - & - \\
\multicolumn{4}{l}{$\bullet$ Lossy Compression:}\\
\rowcolor{Gray}
\hspace{0.5cm}$\rightarrow$ Quantization & \cite{choi2018universal}\cite{choi2016towards} & \cite{alistarh2017qsgd}\cite{wen2017terngrad}\cite{wang2018atomo}\cite{bernstein2018signsgd} & \cite{konevcny2016federated}\cite{caldas2018expanding}\cite{sattler2019robust} & \cite{reisizadeh2019robust}
\cite{koloskova2019decentralized} \\
\hspace{0.5cm}$\rightarrow$ Sparsification & - & \cite{lin2017deep}\cite{aji2017sparse} &\cite{sattler2019robust}\cite{konevcny2016federated}\cite{caldas2018expanding}& \cite{koloskova2019decentralized} \\
\rowcolor{Gray}
\hspace{0.5cm}$\rightarrow$ Sketching & - & \cite{ivkin2019communication}& \cite{li2019privacy} & \cite{ivkin2019communication} \\
\hspace{0.5cm}$\rightarrow$ Low-Rank Approx. & - & \cite{vogels2019powersgd} & \cite{konevcny2016federated} & -\\ 
\rowcolor{Gray}
$\bullet$ Error Accumulation & - & \cite{lin2017deep}\cite{stich2018sparsified}\cite{karimireddy2019error} & \cite{sattler2019sparse} & \cite{tang2019texttt}\\
$\bullet$ Communication Delay & - & \cite{yu2019parallel}\cite{stich2018local}\cite{sattler2019sparse}& \cite{mcmahan2016communication} & \cite{wang2018cooperative}  \\
\rowcolor{Gray}
$\bullet$ Loss-Less Compression & \cite{wiedemann2019deepcabac}\cite{wiedemann2019compact} & \cite{sattler2019sparse}& \cite{sattler2019robust} & - 
\end{tabular}
\end{table*}
\subsection{Federated Learning}
Federated learning \cite{mcmahan2016communication}\cite{li2019federated}\cite{kairouz2019advances} allows multiple parties to jointly train a neural network on their combined data, without having to compromise the privacy of any of the participants. This is achieved by iterating over multiple communication rounds of the following three step protocol:
\begin{itemize}
\item[(1)] The server selects subset of the entire client population to participate in this communication round and communicates a common model initialization to these clients.
\item[(2)] Next, the selected clients compute an update to the model initialization using their private local data. 
\item[(3)] Finally, the participating clients communicate their model updates back to the server where they are aggregated (by e.g.\ an averaging operation) to create a new master model which is used as the initialization point of the next communication round.
\end{itemize}
 Since private data never leaves the local devices, federated learning can provide strong privacy guarantees to the participants. These guarantees can be made rigorous by applying homomorphic encryption to the communicated parameter updates \cite{bonawitz2016practical} or by concealing them with differentially private mechanisms \cite{geyer2017differentially}.

Since in most federated learning applications the training data on a given client is generated based on the specific environment or usage pattern of the sensor, the distribution of data among the clients will usually be ``non-iid'' meaning that any particular user's local dataset will not be representative of the whole distribution. The amount of local data is also typically unbalanced among clients, since different users may make use of their device or a specific application to different extent. Many scenarios are imaginable in which the total number of clients participating in the optimization is much larger than the average number of training data examples per client. The intrinsic heterogeneity of client data in federated learning introduces new challenges when it comes to designing (communication efficient) training algorithms.  

A major issue in federated learning is the massive communication overhead that arises from sending around the model updates. When naively following the federated learning protocol, every participating client has to download and upload a full model during every training iteration. Every such update is of the same size as the full model, which can be in the range of gigabytes for modern architectures with millions of parameters. At the same time, mobile connections are often slow, expensive and unreliable, aggravating the problem further.

\textbf{Compression for Federated Learning:} The most widely used method for reducing communication overhead in federated learning (see Table \ref{tab:1}) is to delay synchronization by letting the clients perform multiple local updates instead of just one \cite{kamp2018efficient}. Experiments show that this way communication can be delayed for up to multiple local epochs without any loss in convergence speed if the clients hold iid data (meaning that all client's data was sampled independently from the same distribution) \cite{mcmahan2016communication}. Communication delay reduces both the downstream communication from the server to the clients and the upstream communication from the clients to the server equally. It also reduces the total number of communication rounds, which is especially beneficial under the constraints of the federated setting as it mitigates the impact of network latency and allows the clients to perform computation off-line and delay communication until a fast network connection is available. 

However, different recent studies show that communication delay drastically slows down convergence in non-iid settings, where the local client’s data distributions are highly divergent \cite{zhao2018federated}\cite{sattler2019robust}. Different methods have been proposed to improve communication delay in the non-iid setting, with varying success: FedProx \cite{sahu2018convergence} limits the divergence of the locally trained models by adding a regularization constraint. Other authors \cite{zhao2018federated} propose mixing in iid public training data with every local client's data. This of course is only possible if such public data is available.
The issue of heterogeneity can also be addressed with Multi-Task and Meta-Learning approaches. First steps towards an adaptive federated learning schemes have been made \cite{sattler2019clustered}\cite{jiang2019improving}, but the heterogeneity issue is still largely unsolved.

Communication delay produces model-updates, which can be compressed further before communication and a variety of techniques have been proposed to this end. In this context it is important to remember the asymmetry between upstream and downstream communication during federated learning: During upstream communication, the server receives model updates from potentially a very large number of clients, which are then aggregated using e.g. an averaging operation. This averaging over the contributions from multiple clients allows for a stronger compression of every individual update. In particular, for unbiased compression techniques it follows directly from the central limit theorem, that the individual upstream updates can be made arbitrarily small, while preserving a fixed error, as long as the number of clients is large enough. Compressing the upstream is also made easier by the fact that the server is always up-to-date with the latest model, which allows the clients to send difference models instead of full models. These difference models contain less information and are thus less sensitive to compression. As clients typically do not participate in every communication round, their local models are often outdated and thus sending difference models is not possible during downstream.

For the above reasons, most existing works on improving communication efficiency in federated learning only focus on the upstream communication (see Table \ref{tab:1}). One line of research confines the parameter update space of the clients to a lower dimensional subspace, by imposing e.g.\ a low-rank or sparsity constraint \cite{konevcny2016federated}. Federated dropout \cite{caldas2018expanding} reduces communication in both upstream and downstream by letting clients train smaller sub-models, which are then assembled into a larger model at the server after every communication round.  As the empirical benefits of training time compression seem to be limited, the majority of methods uses post-hoc compression techniques. Probabilistic quantization and sub-sampling can be used in addition to other techniques such as DeepCABAC \cite{wiedemann2019deepcabac} or sparse binary compression \cite{sattler2019sparse}.

Federated Learning typically assumes a star-shape communication topology, where all clients directly communicate with the server. In some situations it might however be beneficial to consider also hierarchical communication topologies where the devices are organized at multiple levels. This communication topology naturally arises for instance in massively distributed IoT settings, where geographically proximal devices are connected to the same edge server. In these situations, hierarchical aggregation of client contributions can help to reduce the communication overhead by intelligently adapting the communication to the network constraints \cite{liu2019edge}\cite{abad2019hierarchical}.

\subsection{Peer-to-Peer Learning}
Training with one centralized server might be undesirable in some scenarios, because it introduces a single point of failure and requires the clients to trust a centralized entity (at least to a certain degree). Fully decentralized peer-to-peer learning \cite{vanhaesebrouck2017decentralized}\cite{tang2018d}\cite{bellet2017personalized}\cite{lalitha2019peer} overcomes these issues, as it allows clients to directly communicate with one another. In this scenario it is usually assumed that the connectivity structure between the clients is given by a connected graph. Given a certain connectivity structure between the clients, peer-to-peer learning is typically realized via a gossip communication protocol, where in each communication round all clients perform one or multiple steps of stochastic gradient descent and then average their local model with those from all their peers. Communication in peer-to-peer learning may thus be high frequent and involve a large number of clients (see Table \ref{tab:1}). As clients typically are embodied by mobile or IoT devices which collect local data, peer-to-peer learning shares many properties and constraints of federated learning. In particular, the issues related to non-iid data discussed above apply in a similar fashion. A unique characteristic of peer-to-peer learning is that there is no central entity which orchestrates the training process. Making decisions about training related meta parameters may thus require additional consensus mechanisms, which could be realized e.g. via block chain technology \cite{chen2018machine}.

\textbf{Compression for Peer-to-Peer Learning:} Communication efficient peer-to-peer learning of neural networks is a relatively young field of research, and thus the number of proposed compression methods is still limited. However, first promising results have already been achieved with quantization\cite{reisizadeh2019robust}, sketching techniques \cite{ivkin2019communication} and biased compression methods in conjunction with error accumulation \cite{koloskova2019decentralized}\cite{koloskova2019decentralized2}.

\subsection{Distributed Training in the Data Center}
Training modern neural network architectures with millions of parameters on huge datasets such as ImageNet can take prohibitively long time, even on the latest high-end hardware. In distributed training in the data center, the computation of stochastic mini-batch gradients is parallelized over multiple machines in order to reduce training time. In order to keep the compute devices synchronized during this process, they need to communicate their locally communicated gradient updates after every iteration, which results in very high-frequent communication of neural data. This communication is time consuming for large neural network architectures and limits the benefits of parallelization according to Amdahl's law \cite{skillicorn2005foundations}.

\textbf{Compression for Training in the Data-Center:} A large body of research has been devoted to the development of gradient compression techniques. These methods can be roughly organized into two groups: Unbiased and biased compression methods. \emph{Unbiased} (probabilistic) compression methods like QSGD \cite{alistarh2017qsgd}, TernGrad \cite{wen2017terngrad} and \cite{wang2018atomo} reduce the bitwidth of the gradient updates in such a way that the expected quantization error is zero. Since these methods can be easily understood within the framework of stochastic gradient based optimization, establishing convergence is straight forward. However the compression gains achievable with unbiased quantization are limited, which makes these methods unpopular in practice. \emph{Biased} compression methods on the other hand empirically achieve much more aggressive compression rates, at the cost of inflicting a systematic error on the gradients upon quantization, which makes convergence analysis more challenging. An established technique to reduce the impact of biased compression on the convergence speed is error accumulation. In error accumulation the compute nodes keep track of all quantization errors inflicted during training and add the accumulated errors to every newly computed gradient. This way, the gradient information which would otherwise be destroyed by aggressive quantization is retained and carried over to the next iteration. In a key theoretical contribution is was shown \cite{stich2018sparsified}\cite{karimireddy2019error} that the asymptotic convergence rate of SGD is preserved under the application of all compression operators which satisfy a certain contraction property. These compression operators include random sparsification \cite{stich2018sparsified}, top-k sparsification \cite{lin2017deep}, low rank approximations \cite{vogels2019powersgd}, sketching  \cite{ivkin2019communication} and deterministic binarization methods like signSGD \cite{bernstein2018signsgd}. 

All these methods come with different trade-offs with respect to achievable compression rate, computational overhead of encoding and decoding and suitability for different model aggregation schemes. For instance, compression methods based on top-k sparsification with error accumulation \cite{lin2017deep} achieve impressive compression rates of more than $\times 500$ at only marginal loss of convergence speed in terms of training iterations, however these methods also have relatively high computational overhead and do not harmonize well with all-reduce based parameter aggregation protocols \cite{vogels2019powersgd}. 
	
The most typical connectivity structure in distributed training in the data center, is an all-to-all connection topology where all computing devices are directly connected via hard-wire. An all-to-all connection allows for efficient model update aggregation via all-reduce operations \cite{dean2008mapreduce}. However, to efficiently make use of these primitives, compressed representations need to be summable. This property is satisfied for instance by sketches \cite{ivkin2019communication} and low-rank approximations \cite{vogels2019powersgd}.

\section{Related Challenges in Embedded ML}
Despite the recent progress made in efficient deep neural network communication, many unresolved issues still remain.
Some of the most pressing challenges for Embedded ML include:

\textbf{Energy Efficiency:} Since mobile and IoT devices usually have very limited computational resources, Embedded ML solutions are required to be energy efficient. Although many research works aim to reduce the complexity of models through neural architecture search \cite{wu2019fbnet}, design energy-efficient neural network representations \cite{wiedemann2019compact}, or tailor energy-efficient hardware components \cite{chen2016diannao}, the energy efficiency of on-device inference is still a big challenge.

\textbf{Convergence:} An important theoretical concern when designing compression methods for distributed training schemes is that of convergence. While the convergence properties of vanilla stochastic gradient descent based algorithms and many of their distributed variants are well understood \cite{bottou1998online}\cite{kamp2018efficient}\cite{lalitha2019peer}, the assumption of statistical non-iid-ness of the clients data in many Embedded ML applications still pose a set of novel challenges, especially when compression methods are used. 

\textbf{Privacy and Robustness:} Embedded ml applications promise to preserve the privacy of the local datasets. However, multiple recent works have demonstrated that in adversarial settings information about the training data can be leaked via the parameter updates \cite{hitaj2017deep}. A combination of cryptographic techniques such as Secure Multi-Party Computation \cite{goldreich1998secure} and Trusted Execution Environments \cite{subramanyan2017formal}, as well a quantifiable privacy guarantees provided by differential privacy \cite{dwork2014algorithmic} can help to overcome these issues. However it is still unclear how these techniques can be effectively combined with methods for compressed communication and what optimal trade-offs can be made between communication-efficiency and privacy guarantees.

Since privacy guarantees conceal information about the participating clients and their data, there is also an inherent trade-off between privacy and robustness, which needs to be more thoroughly investigated. For instance, it has been shown that it is possible for an adversary to introduce hidden functionality into the jointly trained model \cite{bagdasaryan2018backdoor} or disturb the training process \cite{chen2017distributed}. Detecting these adversarial behaviors becomes much more difficult under privacy constraints. Future methods for data-local training will have to jointly address the issues of efficiency, privacy and robustness.

\textbf{Synchrony:} In most distributed learning schemes of Embedded ML, communication takes place at regular time intervals such that the state of the system can always be uniquely determined \cite{chen2016revisiting}. This has the benefit that it severely simplifies the theoretical analysis of the properties of the distributed learning system. However synchronous schemes may suffer dramatically from delayed computation in the presence of slow workers (stragglers). While countermeasures against stragglers can usually be taken (e.g. by restricting the maximum computation time per worker), in some situations it might still be beneficial to adopt a asynchronous training strategy (e.g. \cite{recht2011hogwild}), where parameter updates are applied to the central model directly after they arrive at the server. This approach avoids delays when the time required by workers to compute parameter updates varies heavily. The absence of a central state however makes convergence analysis far more challenging (although convergence guarantees can still be given \cite{de2015taming}) and may cause model updates to become "stale" \cite{zhang2015staleness}. Since the central model may be updated an arbitrary number of times while a client is computing a model update, this update will often be out-of-date when it arrives at the server. Staleness slows down convergence, especially during the final stages of training.

\textbf{Standards:} To communicate neural data in an interoperable manner, standardized data formats and communication protocols are required. Currently, MPEG is working towards a new part 17 of the ISO/IEC 15938 standard, defining tools for compression of neural data for multimedia applications and representing the resulting bitstreams for efficient transport. Further steps are needed in this direction for a large-scale implementation of Embedded machine learning solutions.

\section{Conclusion}
We currently witness a convergence between the areas of machine learning and communication technology. Not only are today's algorithms used to enhance the design and management of networks and communication components \cite{ibnkahla2000applications}, ML models such as deep neural networks themselves are being communicated more and more in our highly connected world. The roll-out of data-intensive 5G networks and the rise of mobile and IoT applications will further accelerate this development, and it can be predicted that neural data will soon account for a sizable portion of the traffic through global communication networks.

This paper has described the four most important settings in which deep neural networks are communicated and has discussed the respective proposed compression methods and methodological challenges. Our holistic view has revealed that these four seemingly different and independently developing fields of research have a lot in common. We therefore believe that these settings should be considered in conjunction in the future.

\bibliographystyle{abbrv}
\bibliography{sample_large}

\end{document}